\documentclass[runningheads]{llncs}

\usepackage{eccv}

\usepackage{eccvabbrv}

\usepackage{graphicx}
\usepackage{booktabs}

\usepackage{hyperref}

\usepackage{orcidlink}
\usepackage{multirow}

\begin{document}

\title{Rethinking the Good Enough Embedding for Easy Few-Shot Learning} 

\titlerunning{Rethinking the Good Enough Embedding}

% ICPR 2026 is Single-Blind: Authors must be included

\author{Michael Karnes\inst{1} \and
Alper Yilmaz\inst{1}}

\authorrunning{M.~Karnes et al.}

\institute{The Ohio State University, Columbus, OH 43210, USA\\
\email{{karnes.30, yilmaz.15\}@osu.edu}}}

\maketitle

\begin{abstract}
The field of deep visual recognition is undergoing a paradigm shift toward universal representations. The Platonic Representation Hypothesis suggests that diverse architectures trained on massive datasets are converging toward a shared, "ideal" latent space. This again raises a critical question: is a "Good Embedding All You Need?" In this paper, we leverage this convergence to demonstrate that off-the-shelf embeddings are inherently "good enough" for complex tasks, rendering intensive task-specific fine-tuning unnecessary. We explore this hypothesis within the few-shot learning framework, proposing a straightforward, non-parametric pipeline that entirely bypasses backpropagation. By utilizing a $k$-Nearest Neighbor classifier on frozen DINOv2-L features, we conduct a layer-wise characterization to identify an optimal feature extraction. We further demonstrate that manifold refinement via PCA and ICA provides a beneficial regularizing effect. Our results across four major benchmarks demonstrate that our approach consistently surpasses sophisticated meta-learning algorithms, achieving state-of-the-art performance.
\end{abstract}

\section{Introduction}
\label{sec:intro}

Deep visual pattern recognition has been significantly adopted across a wide range of applications. Yet, there still remain significant bottlenecks that limit its implementation. These include the requirement of large datasets, the training time of backpropagation approaches, and the opacity of decisions \cite{He2016,Russakovsky2015,Singh2025}. Few-shot learning (FSL) is designed to address this by measuring a model’s ability to quickly adapt to new environments and tasks using extremely limited training data \cite{rethinking2020,xu2023improving,tang2025}. While humans exhibit an exceptional ability to recognize categories from only a handful of samples, standard deep learning models typically rely on massive annotated datasets to function effectively \cite{visfill2001}.

To tackle this, the research community has largely embraced the meta-learning or “learning to learn” paradigm \cite{maml2017,khadse2025,metaopnet2019,chen2021metabase,rusu2019leo}. This approach defines a family of tasks divided into disjoint meta-training and meta-testing sets, where the learner is evaluated on its average test accuracy across many sampled tasks. Historically, these methods are cast into two categories: optimization-based methods, which design algorithms for rapid task-specific adaptation, and metric-based methods, which learn kernel functions to bypass inner-loop optimization \cite{khadse2025}. However, recent work has sparked a critical debate: is it the complex meta-learning algorithm or the quality of the learned representation that drives fast adaptation? While some suggest feature reuse is the primary factor, others have shown that simple transductive fine-tuning or improved standard fine-tuning models can perform nearly as well as, or even rival, state-of-the-art meta-learning algorithms.

Despite the diversity of these meta-learning paradigms, the reliance on intensive fine-tuning or complex architectures often mirrors the very bottlenecks FSL intends to solve \cite{tang2025,xu2023improving}. Some baselines utilize transductive information from the testing data, which may not always be feasible in real-world scenarios. Furthermore, while visual prior knowledge is essential, there remains a substantial gap between current models and the human ability to utilize semantic and contextual cues. Existing attempts to integrate semantic knowledge, such as word vectors or manual attribute annotations, often suffer from noise or high labor costs, failing to replicate the seamless way humans utilize accumulated experience.

This debate over the primacy of representations is further illuminated by the Platonic Representation Hypothesis \cite{huh2024}. This hypothesis suggests that as AI models grow in scale and are trained on diverse objectives (e.g., vision, text, or multimodal tasks), their internal representations begin to converge toward a shared, "ideal" statistical manifold of reality. If different architectures and training objectives eventually yield a similar underlying structure of the world, it implies that the bottleneck in FSL is not the specific adaptation algorithm, but rather how closely a model's representation aligns with this universal manifold. In this view, a sufficiently powerful pre-trained embedding already contains the necessary structural information to discern novel classes, provided it has reached this convergent state.

In this paper, we propose an exceptionally simple baseline demonstrating that high-quality latent representations are fundamentally more potent for few-shot classification than the current landscape of complex meta-learning algorithms. Our pipeline leverages the frozen latent space of off-the-shelf Deep Neural Networks (DNNs), entirely bypassing the need for backpropagation or task-specific fine-tuning. By systematically isolating the most salient features within the DNN layers, we encode the embedded feature volume into robust representation vectors. This approach facilitates the generation of a concept dictionary and enables k-Nearest Neighbor (kNN) classification. The resulting framework is highly generalizable, rapidly trained, and produces inherently interpretable decision paths.

\section{Related Works}
\label{sec:related_works}
The core idea in metric-based meta-learning is related to nearest neighbor algorithms and kernel density estimation. These methods embed input data into fixed dimensional vectors and use them to design proper kernel functions \cite{khadse2025}. The predicted label of a query is typically the weighted sum of labels over support samples. Early researchers used Siamese networks to encode image pairs and predict confidence scores, while later developments introduced Matching Networks \cite{matching2016} which employed dual networks for query and support samples alongside attention mechanisms. Prototypical Networks \cite{proto2017} learned to encode samples into a shared embedding space where classification is based on the distance to class prototypes. Other variations, such as Relation Networks \cite{relation2018}, leveraged relational modules to represent metrics, while systems like TADAM proposed metric scaling and task conditioning to boost performance \cite{TADAM2018}.

Because deep learning models are neither designed to train with very few examples nor to converge very quickly, optimization-based methods intend to learn specialized adaptation strategies. Some meta-learners exploited memory-augmented architectures like LSTMs to balance the quick acquisition of task-specific knowledge with the slow extraction of transferable features \cite{Santoro2016,ravi2017optimization,systems13070534}. A prominent general optimization algorithm, MAML, aims to find model parameters such that a small number of gradient steps produces large improvements on new tasks \cite{maml2017}. Subsequent iterations simplified this by removing re-initialization requirements or decoupling the optimization from high-dimensional model parameters, such as in the LEO framework which learns a stochastic latent space \cite{rusu2019leo}. Other improvements, like MetaOptNet, replaced linear predictors with differentiable quadratic programming solvers to allow for end-to-end learning with SVMs \cite{metaopnet2019}.

Semantic-based few-shot learning methods overcome the shortcomings of visual approaches by leveraging auxiliary information such as attributes, word embeddings, or knowledge graphs \cite{tang2025,radford2021}. Some models combine label embeddings from class names with visual prototypes to generate adaptive semantic representations \cite{li2025}. Knowledge graphs also provide valuable correlation cues, allowing for the transfer of relationships from base to novel categories via graph convolutional networks \cite{Wang2018,garcia2018fewshot,miller1995}. Nevertheless, these methods often rely on sparse attribute annotations or limited word vectors that may lack contextual richness. The rise of Large Multimodal Models that encapsulate abundant implicit knowledge provides a way to extract and distill higher-quality prior knowledge \cite{zhang2024simple,radford2021}. By leveraging these broad repositories of information, it is possible to significantly boost recognition performance even when using a relatively simple underlying model \cite{tang2025}.

To understand why these strategies work, many efforts have analyzed them through the lenses of optimization and generalization. Research has shown that certain variants work because they converge toward solutions close to each task’s optimal manifold. Critically, studies analyzing whether effectiveness is due to rapid learning or the reuse of high-quality features concluded that feature reuse is the dominant component \cite{rethinking2020,Luo2023}. This finding supports the shift toward visual-based methods that transfer prior knowledge from base classes to novel classes by pre-training powerful feature extractors and directly generating classifier weights.

\section{Methods}
\label{sec:related_works}
\subsection{Problem Formulation}

In the few-shot learning paradigm, we define a distribution of tasks $T$ from which individual tasks are sampled. The goal is to develop a model that can generalize across these tasks despite having access to only a limited number of training examples per task.

\subsection{The Meta-Learning Framework}
The meta-training set is defined as a collection of tasks $\mathcal{T} = \{(D_{train, i}, D_{test, i})\}_{i=1}^I$, often termed the meta-training set. Each task consists of a small training (support) set $D_{train} = \{(x_t, y_t)\}_{t=1}^T$ and a testing (query) set $D_{test} = \{(x_q, y_q)\}_{q=1}^Q$, both sampled from the same distribution. 

The objective of FSL visual classification is to generate a function that predicts the class of a query sample image given a set of reference data. To formalize this, we assume a dataset of $N$ images, $D = \{(x_i, y_i)\}_{i=1}^N$, where each image $x_i \in \mathbb{R}^{d_{img}}$ is associated with a label $y_i \in \{1, \dots, C\}$. 

The FSL approach utilizes a sub-sample $D^{\tau} \subseteq D$ to construct two distinct sets for a given task:
\begin{itemize}
    \item \textbf{Support Set ($S^{\tau}$):} A set of reference images $S^{\tau} = \{(x_i, y_i)\}_{i=1}^{N^{\tau}}$ used to generate encoded appearance models of novel classes (a concept dictionary).
    \item \textbf{Query Set ($Q^{\tau}$):} A set of unlabeled images $Q^{\tau} = \{(x_i^*, y_i^*)\}_{i=1}^{N^*}$ that the system must accurately classify based on its relation to the support set.
\end{itemize}

In practice, several constraints are applied to these sets to ensure a valid few-shot scenario. $D^{\tau}$ must consist of a sub-sample of classes, each containing a strictly specified number of samples, $k$, commonly referred to as $k$-shot learning. Furthermore, $S^{\tau}$ and $Q^{\tau}$ are split such that any class present in the query set is also represented within the support set to allow for proper metric comparison. 

The primary FSL objective is to find an optimal function $f(S^{\tau}, Q^{\tau})$ that provides the most accurate estimation of the query set labels $y^*$. This best estimate is mathematically defined by maximizing the expected local likelihood across different task instances $\tau$:

\begin{equation}
\mathbb{E}_{\tau} \left[ \prod_{Q^*} p(y_i^* | f(x_i^*, S^{\tau})) \right]
\end{equation}

\subsection{Meta-Training and Evaluation}
The overarching objective of meta-learning algorithms is to learn a high-quality embedding model $\phi$, such that the average test error of the base learner across a distribution of tasks is minimized. Formally, this is expressed as:

\begin{equation}
\phi^* = \arg\min_{\phi} E_{\mathcal{T}}[L_{meta}(D_{test}; \phi, \omega)] \quad \text{where} \quad \omega = A(D_{train}; \phi)
\end{equation}

Once the meta-training phase is finished, the performance of the model is evaluated on a set of held-out tasks $S = \{(D_{train, j}, D_{test, j})\}_{j=1}^J$, called the meta-testing set. The evaluation is conducted over the distribution of the test tasks:

\begin{equation}
E_{S}[L_{meta}(D_{test}; \phi^*, \omega)] \quad \text{where} \quad \omega = A(D_{train}; \phi^*)
\end{equation}

\subsection{Theoretical Foundation: The Latent Manifold as a GMM}
Deep neural image classification networks are traditionally trained to estimate the class probabilities of a given image via a final $softmax()$ output layer. This iterative regression process generates learned feature-extracting filters that correspond to the most discriminative attributes of the training set. Assuming these filters are sufficiently generalized, a query image $x^*$ can be effectively encoded within the network's latent feature space $\Phi$:

\begin{equation}
p(y^*=c|x^*) = softmax(\Phi(x^*))
\end{equation}

These class probabilities can also be interpreted as a Gaussian Mixture Model (GMM) of class characteristics residing in the latent manifold \cite{bateni2020improved}. Under this framework, an input image is treated as a specific instance of visual characteristics produced by a GMM source. 

Because the DNN performs a hierarchical series of linear kernel transformations, the Central Limit Theorem suggests that the linear combination of these distributions, or the partial sum of randomly indexed m-dependent variables, tends toward a Gaussian distribution. Consequently, the DNN-embedded features can be viewed as a GMM produced from the underlying visual characteristics, where the class probability is defined by:

\begin{equation}
p(y^*=c | x^*) = \frac{\pi_c \mathcal{N}(x^*; \mu_c, \Sigma_c)}{\sum_{c'} \pi_{c'} \mathcal{N}(x^*; \mu_{c'}, \Sigma_{c'})}
\end{equation}

\subsection{Feature Extraction and Classification Process}
\begin{figure}[t]
\centering\begin{tikzpicture}[node distance=1.2cm, font=\sffamily\scriptsize,
block/.style={rectangle, draw, fill=blue!5, text width=1.8cm, text centered, minimum height=0.8cm},
special/.style={rectangle, draw, fill=orange!10, text width=1.8cm, text centered, minimum height=0.8cm},
dict/.style={rectangle, draw, fill=green!5, text width=1.8cm, text centered, minimum height=1cm}]
\node (in) [block] {Input Image};
\node (vit) [block, right=of in, xshift=-0.1cm] {ViT Encoder Block (L)};
\node (layer) [special, right=of vit, xshift=-0.1cm] {Extract Block (L) Features};
\node (gap) [block, right=of layer, xshift=-0.1cm] {Global Average Pooling};
\node (pca) [block, below=of gap] {PCA/ICA Reduction};
\node (dict) [dict, left=of pca, xshift=0.1cm] {Concept Dictionary {$\mu_c, \Sigma_c$}};
\node (maha) [block, left=of dict, xshift=0.1cm] {Mahalanobis Distance};
\node (nn) [block, left=of maha, xshift=0.1cm, fill=red!5] {Nearest Neighbor};
\draw [-Stealth, thick] (in) -- (vit);
\draw [-Stealth, thick] (vit) -- (layer);
\draw [-Stealth, thick] (layer) -- (gap);
\draw [-Stealth, thick] (gap) -- (pca);
\draw [-Stealth, thick] (pca) -- (dict);
\draw [-Stealth, thick] (dict) -- (maha);
\draw [-Stealth, thick] (maha) -- (nn);
\end{tikzpicture}
\caption{The proposed feature extraction and classification pipeline.}
\label{fig:pipeline}
\end{figure}
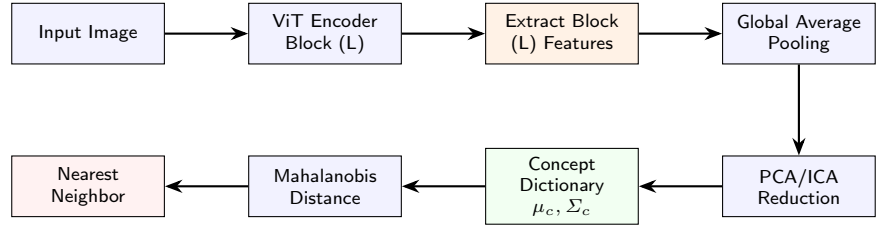

The full pipeline from extraction to classification is visualized in Figure \ref{fig:pipeline}.
\subsubsection{Stage 1: Unsupervised Encoding of Platonic Ideals}
 The framework first derives a dimensionality reduction transform from unlabeled images by extracting latent features from a specific transformer block of a frozen DINOv2-Large backbone \cite{oquab2024dinov}. For an input $x$, the global latent vector $z \in \mathbb{R}^{1024}$ is obtained by averaging patch tokens and cls token $N=257$:

\begin{equation}
z = \frac{1}{N} \sum_{i=1}^{N} p_{i,l}
\end{equation}

where $p_{i,l}$ represents the $i$-th patch token. The Principal Component Analysis (PCA) or Independent Component Analysis (ICA) projection is then trained from this set of embedded unlabeled images. The PCA or ICA is then used to project the centered latent vector into a reduced space $s \in \mathbb{R}^{d'}$ via:

\begin{equation}
s = W^{T}(z - \mu)
\end{equation}

\subsubsection{Stage 2: Supervised Synthesis of Aristotelian Concepts}
Then the support set of $k$ labeled training samples for each class $c$ is encoded and aggregated into the collection $\mathcal{S}_c = \{s_{1,c}, \dots, s_{n,c}\}$. The covariance $\tilde{\Sigma}_c$ is then calculated for each class $c$.

\subsubsection{Stage 3: Inference via Per-Class Mahalanobis Distance}

During inference, a test image $x_{test}$ is encoded into $s_{test}$. The sample is then evaluated against every exemplar $e_c$ in the dictionary using the Mahalanobis metric, normalized by the projected class covariance $\tilde{\Sigma}_c$:

\begin{equation}
D_M(\tilde{s}_{test}, \tilde{e}_c) = \sqrt{(\tilde{s}_{test} - \tilde{e}_c)^T \tilde{\Sigma}_c^{-1} (\tilde{s}_{test} - \tilde{e}_c)}
\end{equation}

The predicted label is assigned via a majority vote among the $k$ neighbors.

\section{Experiments}
\subsection{Setup}
Our experimental evaluation proceeded in two phases to validate the strength of off-the-shelf representations. First, we conducted a network characterization to map accuracy trends across the hierarchical layers of the frozen backbone, identifying the optimal depth for feature extraction. Second, we transitioned to a few-shot learning study, comparing our non-parametric approach against state-of-the-art benchmarks to demonstrate the effectiveness of fixed features. This methodology builds on the insight provided by the Platonic Representation Hypothesis, which posits that DNNs are all converging toward a common, universal, and highly generalizable latent space embedding.

\subsubsection{Network Characterization}
To identify the most effective source of universal features, we first conducted a systematic layer-wise characterization of the DINOv2-L backbone \cite{oquab2024dinov}. We evaluated the raw latent representations, defined as the feature vectors prior to any dimensionality reduction, across all layers of the network to determine where the most discriminative information, the good embedding, resides. The resulting accuracy-per-layer data were then fitted to a logistic curve to quantify the mathematical relationship between network depth and semantic knowledge extraction.

For this study, we utilized the training splits of four widely used benchmarks: miniImageNet \cite{SunLCS2019MTL,LiuMTLDownload,Russakovsky2015}, tieredImageNet \cite{SunLCS2019MTL,LiuMTLDownload,Russakovsky2015}, FC100 \cite{SunLCS2019MTL,LiuMTLDownload,Krizhevsky09}, and CIFAR-FS \cite{zenodo_cifar100,Krizhevsky09}. We ensured strict data integrity by using only the training data and verifying that there was no overlap with evaluation sets. For each class, we built a support set using 64 labeled images and evaluated the resulting classification performance on 300 query images, all resized to 224 by 224. A $k$ of 15 was used for classification. To maintain computational efficiency while ensuring statistical significance, we performed many-way classification across all available training classes for most datasets. In the case of tieredImageNet, which contains a larger set of 351 training classes, we randomly sub-sampled 100 classes for our characterization. This evaluation allowed us to observe the inherent structure of the latent manifold across the entire depth of the pre-trained model.

After identifying the top-performing layer for each dataset, we investigated the relationship between feature density and classification performance. We repeated the characterization process by applying PCA to the best layer and its immediate neighbors, exploring one layer above and below, or the two preceding layers in the case of an edge-layer selection. By projecting the embedded vectors into reduced subspaces of 512, 256, 128, and 64 components, we aimed to measure the accuracy cost associated with increased representational efficiency. 

\subsubsection{Few-Shot Comparison}
Building upon our characterization results, we evaluated the system’s performance in a standard 5-way few-shot learning setting. Using the optimal layer identified in the previous phase, we tested the classification accuracy for both 1-shot and 5-shot scenarios across all four datasets. To assess the trade-off between representational density and few-shot generalization, we compared the raw high-dimensional features against PCA and ICA reduced embeddings of 512, 256, and 128 components. This alignment with the idea that a "good embedding is all you need" allowed us to verify if reduced universal features remain competitive with higher-dimensional counterparts.

To bolster the limited support data, we applied a lightweight augmentation strategy by generating four additional variants for each reference image using a 50\% horizontal flip probability and 0.1\% additive pixel noise. During inference, we performed classification using $k=5$ nearest neighbors. To ensure statistical stability, we report the mean accuracy and confidence intervals calculated across 600 independent trials for each experimental configuration.

\subsection{Results}
\subsubsection{Network Characterization}
\begin{figure}[htb]
  \centering
  \includegraphics[width=0.9\linewidth]{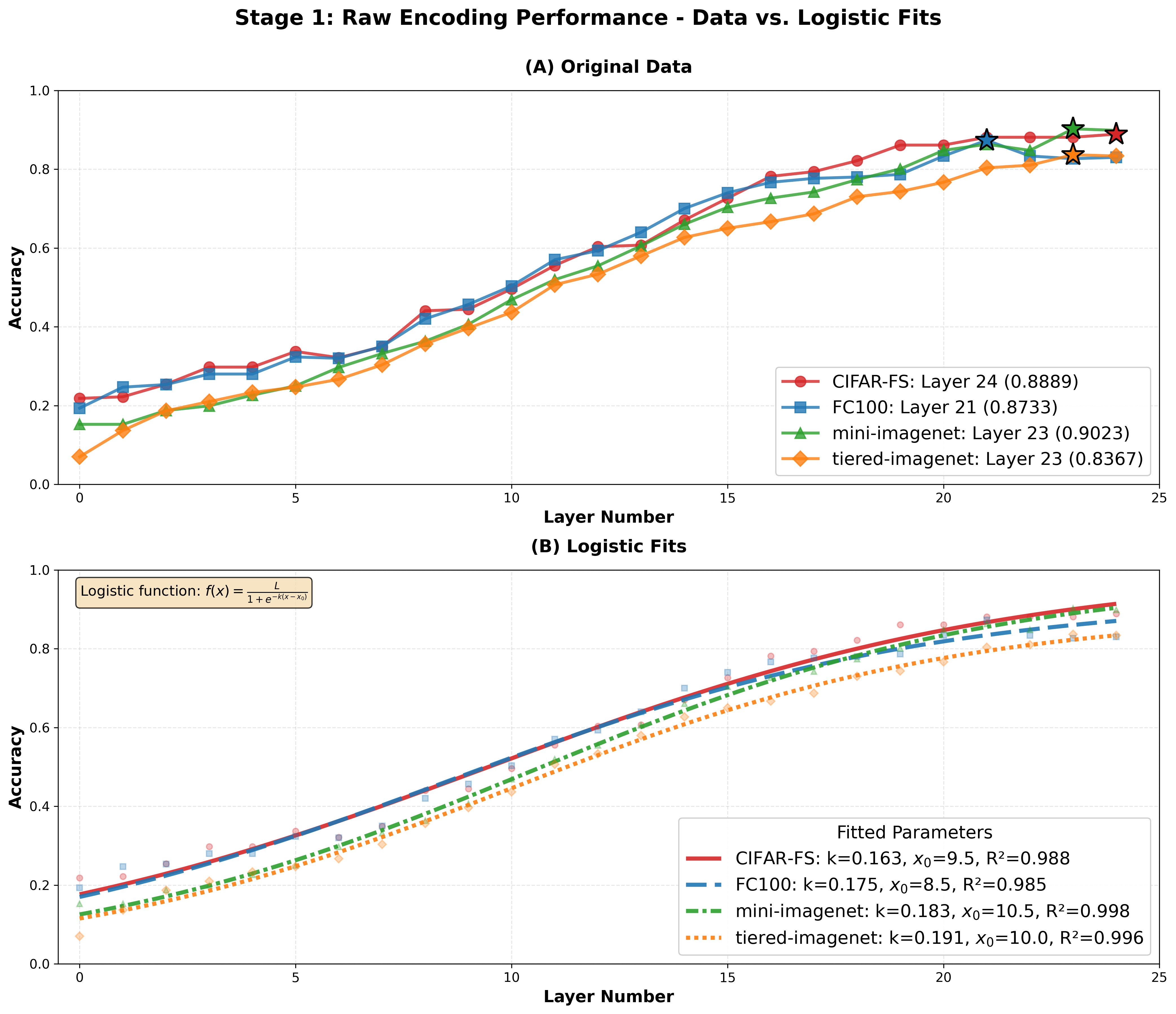}
  \caption{ Layer-wise Accuracy Trends and Logistic Fits. (A) Raw classification performance across the layers of the DINOv2-L backbone. Stars indicate the peak accuracy layer for each dataset. (B) Logistic function fits $f(x) = \frac{L}{1 + e^{-k(x-x_0)}}$ applied to the accuracy data, where the high $R^2$ values ($>0.98$) suggest that feature maturation follows a sigmoidal progression toward a semantic plateau.
  }
  \label{fig:layer_effects}
\end{figure}

We present the results of our layer-wise analysis through two primary lenses: hierarchical layer maturity and the impact of dimensionality reduction. Figure \ref{fig:layer_effects}(A) illustrates the classification accuracy across all the layers of the DINOv2-L backbone \cite{oquab2024dinov}. We observe a consistent upward trend in accuracy as features move toward the deeper layers of the network, with peak performance (marked by stars) occurring between layers 21 and 24 (layer 24 being norm) across all datasets.

To better understand the underlying trend of feature maturation, we applied a logistic curve fit to the data, as shown in Figure \ref{fig:layer_effects}(B). The high $R^2$ values (ranging from 0.985 to 0.998) indicate that the accumulation of semantic knowledge in the latent space follows a predictable sigmoidal progression allowing us to quantify the trade-off between layer depth and classification accuracy. This logistic behavior suggests that there is a rapid learning phase in the middle blocks followed by a saturation of semantic information in the final layers.

\begin{figure}[htb]
  \centering
  \includegraphics[width=0.9\linewidth]{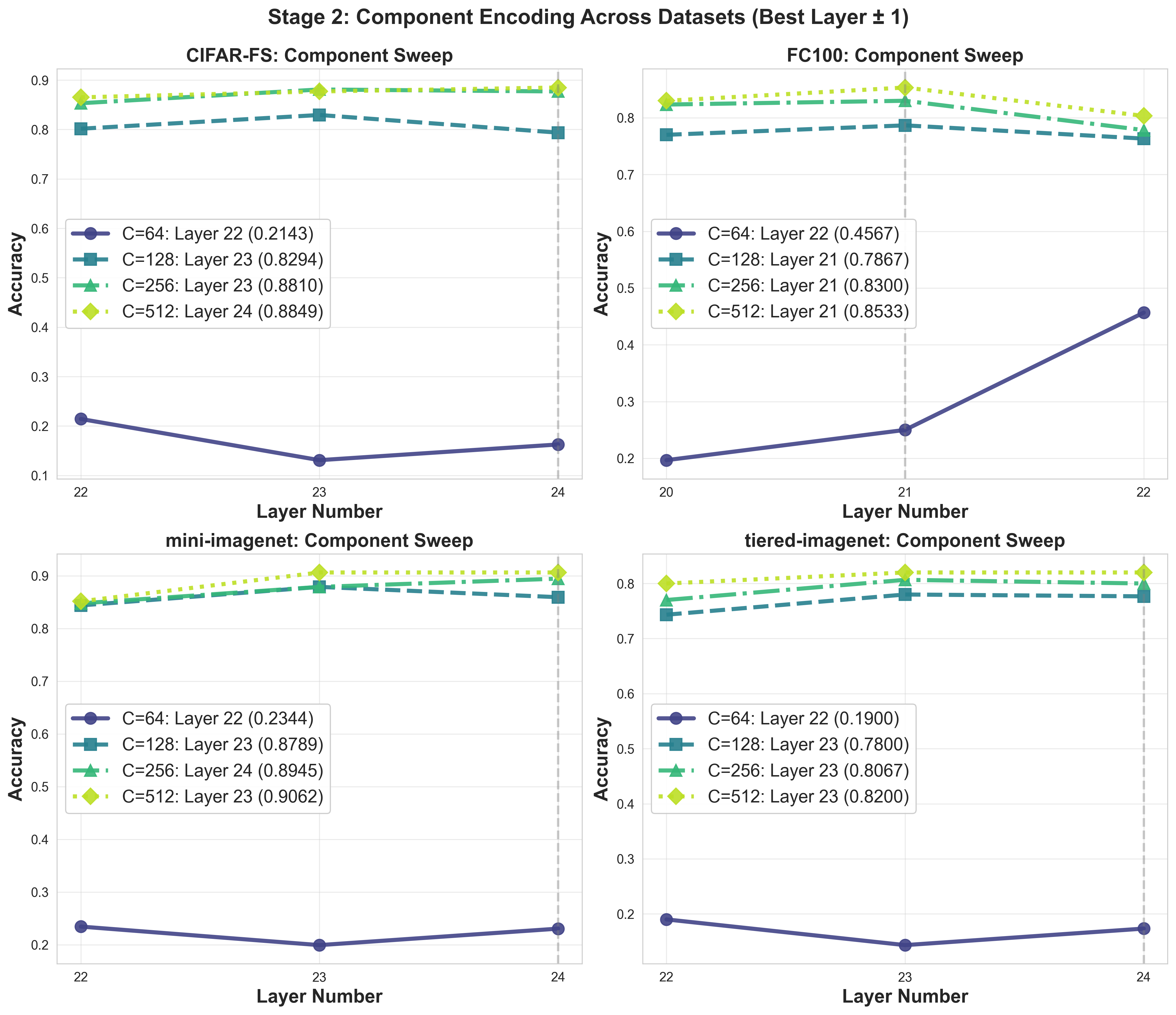}
  \caption{Dimensionality Reduction Resilience Across Latent Layers. Classification accuracy was evaluated across the identified optimal layers and their immediate neighbors using PCA-compressed embeddings. The observed stability suggests that high-quality representations can be significantly condensed for computational efficiency without compromising the "good enough" nature of the embedding for few-shot tasks.
  }
  \label{fig:pca_effects}
\end{figure}

Figure \ref{fig:pca_effects} illustrates the resilience of the identified optimal layers and their immediate neighbors under varying levels of PCA-driven compression. By projecting the latent spaces into 512, 256, 128, and 64 components, we quantify the trade-off between representational efficiency and classification precision. Across all four benchmarks, we observe that accuracy remains remarkably stable when reducing the dimensionality to 512 or 256 components, often maintaining within 1–4\% of the raw high-dimensional performance. Even at 128 components, the embeddings retain significant discriminative power. However, a sharp performance cliff appears at 64 components, where accuracy drops precipitously (often below 30\%), indicating a fundamental limit to the compression of the universal manifold before essential semantic relationships are lost. This robustness at moderate compression levels further establishes that off-the-shelf embeddings are inherently "good enough" to be used as efficient, low-dimensional descriptors for downstream few-shot tasks.

\subsubsection{Few-Shot Comparison}
To conclude our evaluation, we benchmark our framework against established meta-learning algorithms. Having identified the optimal latent layers and demonstrated the resilience of these features to dimensionality reduction, we now apply this methodology to the standard 5-way few-shot classification task. We compare our results utilizing raw high-dimensional features, as well as PCA-compressed and ICA-compressed embeddings, against the state-of-the-art results reported in recent literature \cite{tang2025}, including optimization-based methods like MAML \cite{maml2017} and MetaOptNet \cite{metaopnet2019}, and semantic fusion models such as SYNTRANS \cite{tang2025}. This comparison is divided into two parts: Table \ref{tab:fsl_imgnet_results} focuses on the ImageNet derivatives, miniImageNet and tieredImageNet, while Table \ref{tab:cifar_fc100_results} evaluates erformance on the CIFAR-based benchmarks, CIFAR-FS and FC100. These comparisons serve to validate our core hypothesis: that the robust, universal embedding features of today's large DNN's, when paired with a simple kNN classifier can match or surpass the performance of complex paradigms.

\setlength{\tabcolsep}{1pt}
\begin{table*}[htb]
\centering
\caption{Comparative results on MiniImageNet and TieredImageNet datasets. Average accuracy (\%) with 95\% confidence intervals.}
\label{tab:fsl_imgnet_results}
\begin{scriptsize}
\begin{tabular}{llcccc}
\toprule
\multirow{2}{*}{\textbf{Method}} & \multirow{2}{*}{\textbf{Backbone}} & \multicolumn{2}{c}{\textbf{MiniImageNet}} & \multicolumn{2}{c}{\textbf{TieredImageNet}} \\ \cmidrule(r){3-4} \cmidrule(l){5-6}
 &  & \textbf{5-way 1-shot} & \textbf{5-way 5-shot} & \textbf{5-way 1-shot} & \textbf{5-way 5-shot} \\ \midrule

MatchNet \cite{matching2016} & ResNet-12 & 65.64 $\pm$ 0.20 & 78.72 $\pm$ 0.15 & 68.50 $\pm$ 0.92 & 80.60 $\pm$ 0.71 \\
ProtoNet \cite{proto2017} & ResNet-12 & 62.29 $\pm$ 0.33 & 79.46 $\pm$ 0.48 & 68.25 $\pm$ 0.23 & 84.01 $\pm$ 0.56 \\
MAML \cite{maml2017} & ResNet-12 & 58.05 $\pm$ 0.21 & 58.05 $\pm$ 0.10 & 67.92 $\pm$ 0.17 & 72.41 $\pm$ 0.20 \\
MetaOptNet \cite{metaopnet2019} & ResNet-18 & 62.64 $\pm$ 0.61 & 78.63 $\pm$ 0.46 & 65.99 $\pm$ 0.72 & 81.56 $\pm$ 0.53 \\
RFS \cite{rethinking2020} & SEResNet-12 & 67.73 $\pm$ 0.63 & 83.35 $\pm$ 0.41 & 72.55 $\pm$ 0.69  & 86.72 $\pm$ 0.49 \\ 
FEAT \cite{ye2021} & ResNet-12 & 66.78 $\pm$ 0.20 & 82.05 $\pm$ 0.14 & 70.80 $\pm$ 0.23 & 84.79 $\pm$ 0.16 \\
Meta-Baseline \cite{chen2021metabase} & ResNet-12 & 63.17 $\pm$ 0.23 & 79.26 $\pm$ 0.17 & 68.62 $\pm$ 0.27 & 83.29 $\pm$ 0.18 \\
CVET \cite{yang2022} & ResNet-12 & 70.19 $\pm$ 0.46 & 84.66 $\pm$ 0.29 & 72.62 $\pm$ 0.51 & 86.62 $\pm$ 0.33 \\
FGFL \cite{cheng2023} & ResNet-12 & 69.14 $\pm$ 0.80 & 86.01 $\pm$ 0.62 & 73.21 $\pm$ 0.88 & 87.21 $\pm$ 0.61 \\
SUN \cite{dong2022} & ViT-S & 67.80 $\pm$ 0.45 & 83.25 $\pm$ 0.30 & 72.99 $\pm$ 0.50 & 86.74 $\pm$ 0.33 \\
SMKD \cite{lin2023} & ViT-S & 74.28 $\pm$ 0.18 & 88.82 $\pm$ 0.09 & 78.83 $\pm$ 0.20 & 91.02 $\pm$ 0.12 \\
FewTURE \cite{hiller2022} & Swin-T & 72.40 $\pm$ 0.78 & 86.38 $\pm$ 0.49 & 76.32 $\pm$ 0.87 & 89.96 $\pm$ 0.55 \\ \midrule

KTN \cite{peng2019} & ResNet-12 & 61.42 $\pm$ 0.72 & 70.19 $\pm$ 0.62 & 68.01 $\pm$ 0.73 & 79.06 $\pm$ 0.70 \\
AM3 \cite{xing209} & ResNet-12 & 65.30 $\pm$ 0.49 & 78.10 $\pm$ 0.36 & 69.08 $\pm$ 0.47 & 82.58 $\pm$ 0.31 \\
PC-FSL \cite{zhang2021} & ResNet-12 & 69.68 $\pm$ 0.76 & 81.65 $\pm$ 0.54 & 74.19 $\pm$ 0.90 & 86.09 $\pm$ 0.60 \\
SEGA \cite{yang2021sega} & ResNet-12 & 69.04 $\pm$ 0.26 & 79.03 $\pm$ 0.18 & 72.18 $\pm$ 0.30 & 84.28 $\pm$ 0.21 \\
LPE-CLIP \cite{yang2023Sem} & ResNet-12 & 71.64 $\pm$ 0.40 & 79.67 $\pm$ 0.32 & 73.88 $\pm$ 0.48 & 84.88 $\pm$ 0.36 \\
KSTNET \cite{li2025} & ResNet-12 & 71.51 $\pm$ 0.73 & 82.61 $\pm$ 0.48 & 75.52 $\pm$ 0.77 & 85.85 $\pm$ 0.59 \\
4S-FSL \cite{lu2023} & ResNet-12 & 72.64 $\pm$ 0.70 & 84.73 $\pm$ 0.50 & - & - \\
SP-CLIP \cite{chen2023sem} & Visformer-T & 72.31 $\pm$ 0.40 & 83.42 $\pm$ 0.30 & 78.03 $\pm$ 0.46 & 88.55 $\pm$ 0.32 \\
SemFew \cite{zhang2024simple} & Swin-T & 78.94 $\pm$ 0.66 & 86.49 $\pm$ 0.50 & 82.37 $\pm$ 0.77 & 89.89 $\pm$ 0.52 \\
SYNTRANS \cite{tang2025}& ResNet-12 & 76.20 $\pm$ 0.69 & 86.12 $\pm$ 0.54 & 79.69 $\pm$ 0.81 & 87.78 $\pm$ 0.60 \\
SYNTRANS \cite{tang2025} & ViT-S & 81.30 $\pm$ 0.61 & 89.96 $\pm$ 0.42 & \textbf{84.31} $\pm$ 0.54 & 91.73 $\pm$ 0.44 \\ \midrule

Ours-Raw & DINOv2-L & 83.36 $\pm$ 0.75 & \textbf{\underline{96.51}} $\pm$ 0.24 & 83.61 $\pm$ 0.84 & \textbf{\underline{94.96}} $\pm$ 0.40 \\ \midrule
Ours-PCA 512 & DINOv2-L & 84.11 $\pm$ 0.74 & 96.18 $\pm$ 0.26 & \underline{84.01} $\pm$ 0.81 & 94.50 $\pm$ 0.42 \\
Ours-PCA 256 & DINOv2-L & \textbf{\underline{84.16}} $\pm$ 0.78 & 95.34 $\pm$ 0.31 & 82.23 $\pm$ 0.81 & 93.09 $\pm$ 0.46 \\
Ours-PCA 128 & DINOv2-L & 82.72 $\pm$ 0.83 & 93.94 $\pm$ 0.40 & 79.89 $\pm$ 0.86 & 91.55 $\pm$ 0.49 \\ \midrule

Ours-ICA 512 & DINOv2-L & 72.11 $\pm$ 0.92 & 94.86 $\pm$ 0.27 & 75.05 $\pm$ 1.05 & 94.68 $\pm$ 0.38 \\
Ours-ICA 256 & DINOv2-L & 76.02 $\pm$ 0.86 & 95.40 $\pm$ 0.26 & 79.24 $\pm$ 0.93 & 94.39 $\pm$ 0.41 \\
Ours-ICA 128 & DINOv2-L & 78.24 $\pm$ 0.83 & 94.39 $\pm$ 0.34 & 79.35 $\pm$ 0.86 & 92.93 $\pm$ 0.45 \\
\bottomrule
\end{tabular}
\end{scriptsize}
\end{table*}

We evaluate our proposed method against a broad range of state-of-the-art FSL approaches on the miniImageNet and tieredImageNet datasets, as summarized in Table~\ref{tab:fsl_imgnet_results}. Our approach significantly outperforms existing methods in all cases, with our Raw variant achieving a new state-of-the-art 5-shot accuracy of 96.51\% on miniImageNet and 94.96\% on tieredImageNet, surpassing the previous best results from SYNTRANS \cite{tang2025} by over 6.5\% and 3.2\%, respectively. Furthermore, our results demonstrate that PCA can effectively regularize the latent space, as evidenced by our PCA-256 achieving the overall best 1-shot accuracy on miniImageNet (84.16\%). While ICA variants maintain competitive 5-shot performance, PCA-based proves superior for 1-shot classification.

\begin{table*}[htb]
\centering
\caption{Few-shot classification performance on CIFAR-FS and FC100 datasets. Results report average accuracy (\%) with 95\% confidence intervals.}
\label{tab:cifar_fc100_results}
\begin{scriptsize}
\begin{tabular}{llcccc}
\toprule
\multirow{2}{*}{\textbf{Method}} & \multirow{2}{*}{\textbf{Backbone}} & \multicolumn{2}{c}{\textbf{CIFAR-FS}} & \multicolumn{2}{c}{\textbf{FC100}} \\ \cmidrule(r){3-4} \cmidrule(l){5-6}
 &  & \textbf{5-way 1-shot} & \textbf{5-way 5-shot} & \textbf{5-way 1-shot} & \textbf{5-way 5-shot} \\ \midrule
ProtoNet \cite{proto2017} & ResNet-12 & 72.20 $\pm$ 0.73 & 83.50 $\pm$ 0.50 & 41.54 $\pm$ 0.76 & 57.08 $\pm$ 0.76 \\
MetaOptNet \cite{metaopnet2019} & ResNet-12 & 72.80 $\pm$ 0.70 & 84.30 $\pm$ 0.50 & 47.20 $\pm$ 0.60 & 55.50 $\pm$ 0.60 \\
RFS \cite{rethinking2020} & SEResNet-12 & 75.60 $\pm$ 0.80 & 88.20 $\pm$ 0.50 & 52.00 $\pm$ 0.70 & 68.80 $\pm$ 0.60 \\ 
SUN \cite{dong2022} & ViT-S & 78.37 $\pm$ 0.46 & 88.84 $\pm$ 0.32 & - & - \\
SMKD \cite{lin2023} & ViT-S & 80.08 $\pm$ 0.18 & 90.63 $\pm$ 0.13 & 50.38 $\pm$ 0.16 & 68.37 $\pm$ 0.16 \\
FewTURE \cite{hiller2022} & Swin-T & 77.76 $\pm$ 0.81 & 88.90 $\pm$ 0.59 & 47.68 $\pm$ 0.78 & 63.81 $\pm$ 0.75 \\
SEGA \cite{yang2023Sem} & ResNet-12 & 78.45 $\pm$ 0.24 & 86.00 $\pm$ 0.20 & - & - \\
LPE-CLIP \cite{yang2023Sem} & ResNet-12 & 80.62 $\pm$ 0.41 & 86.22 $\pm$ 0.33 & - & - \\
4S-FSL \cite{lu2023} & ResNet-12 & 74.50 $\pm$ 0.84 & 88.76 $\pm$ 0.53 & - & - \\
SP-CLIP \cite{chen2023sem} & Visformer-T & 82.18 $\pm$ 0.40 & 88.24 $\pm$ 0.32 & 48.53 $\pm$ 0.38 & 61.55 $\pm$ 0.41 \\
SemFew \cite{zhang2024simple} & Swin-T & 84.34 $\pm$ 0.67 & 89.11 $\pm$ 0.54 & 54.27 $\pm$ 0.77 & 65.02 $\pm$ 0.72 \\
SYNTRANS \cite{tang2025} & ResNet-12 & 82.58 $\pm$ 0.75 & 89.42 $\pm$ 0.56 & 52.30 $\pm$ 0.75 & 64.91 $\pm$ 0.59 \\
SYNTRANS \cite{tang2025} & ViT-S & 84.64 $\pm$ 0.65 & 90.81 $\pm$ 0.41 & \textbf{56.38} $\pm$ 0.69 & 69.45 $\pm$ 0.54 \\ \midrule

Ours-Raw & DINOv2-L & 90.92 $\pm$ 0.58 & \textbf{\underline{97.66}} $\pm$ 0.19 & \underline{56.01} $\pm$ 0.92 & 77.67 $\pm$ 0.71 \\ \midrule
Ours-PCA 512 & DINOv2-L & \textbf{\underline{91.35}} $\pm$ 0.57 & 97.53 $\pm$ 0.20 & 55.34 $\pm$ 0.89 & 76.00 $\pm$ 0.71 \\
Ours-PCA 256 & DINOv2-L & 90.71 $\pm$ 0.61 & 97.14 $\pm$ 0.23 & 53.56 $\pm$ 0.86 & 73.20 $\pm$ 0.72 \\
Ours-PCA 128 & DINOv2-L & 89.59 $\pm$ 0.64 & 96.45 $\pm$ 0.28 & 51.26 $\pm$ 0.85 & 69.68 $\pm$ 0.75 \\ \midrule

Ours-ICA 512 & DINOv2-L & 75.19 $\pm$ 0.92 & 95.56 $\pm$ 0.25 & 53.98 $\pm$ 0.92 & \textbf{\underline{78.74}} $\pm$ 0.67 \\
Ours-ICA 256 & DINOv2-L & 81.76 $\pm$ 0.78 & 96.71 $\pm$ 0.21 & 54.79 $\pm$ 0.91 & 76.67 $\pm$ 0.70 \\
Ours-ICA 128 & DINOv2-L & 85.59 $\pm$ 0.69 & 96.53 $\pm$ 0.24 & 52.02 $\pm$ 0.86 & 71.20 $\pm$ 0.76 \\
\bottomrule
\end{tabular}
\end{scriptsize}
\end{table*}
Table \ref{tab:cifar_fc100_results} presents the performance comparison on the CIFAR-FS and FC100 datasets. On CIFAR-FS, our methods establish a new performance ceiling, with our PCA-512 achieving 91.35\% in the 1-shot setting and our Raw reaching 97.66\% in the 5-shot setting, representing a substantial improvement over the previous state-of-the-art. For the more challenging FC100 dataset, while SYNTRANS \cite{tang2025} maintains a slight lead in 1-shot accuracy, our ICA-512 variant achieves the overall best 5-shot accuracy of 78.74\%, outperforming the closest competitor by over 9\%. Notably, the superior performance of ICA on FC100 5-shot shows that it can be advantageous in certain scenarios. Similar to our observations on ImageNet, PCA-based refinement remains highly effective for 1-shot CIFAR-FS classification, further validating low-rank manifold projections.

\begin{figure}[htb]
  \centering
  \includegraphics[width=0.7\linewidth]{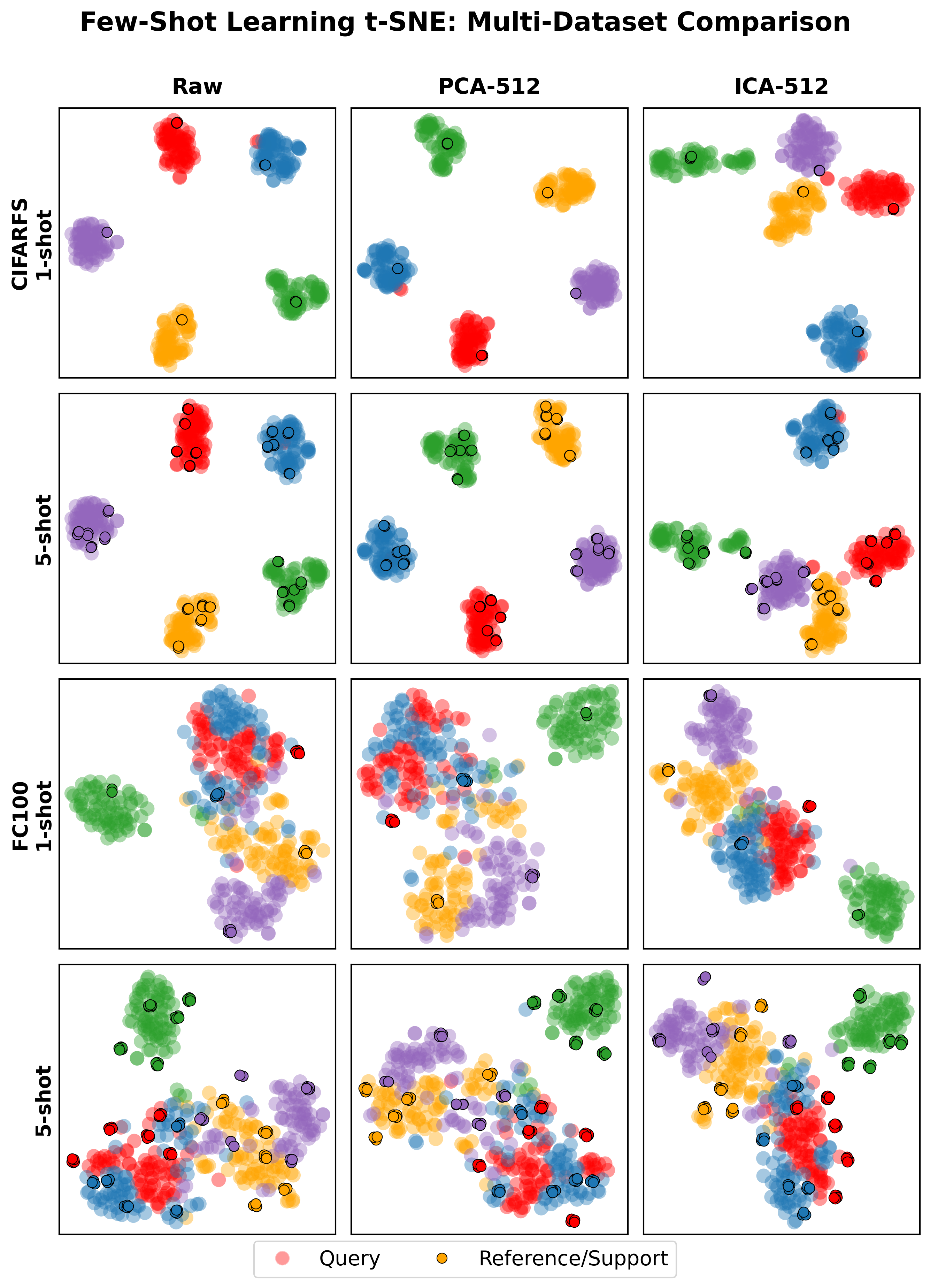}
  \caption{t-SNE visualizations of the latent manifolds for 5-way tasks on the CIFAR-FS (top) and FC100 (bottom) datasets. The figure compares raw high-dimensional features (left) against PCA-512 (center) and ICA-512 (right) dimensionally reduced features. Rows 1 and 3 illustrate the 1-shot scenario, while Rows 2 and 4 represent the 5-shot scenario. Note the increased cluster cohesion in CIFAR-FS versus the higher semantic overlap in the FC100 manifold, aligning with our global accuracy results.
  }
  \label{fig:tsne}
\end{figure}

Figure \ref{fig:tsne} qualitatively evaluates the CIFAR-FS and FC100 latent manifolds generated by our framework across raw, PCA, and ICA representations. For CIFAR-FS, both the raw and PCA-512 embeddings maintained distinct, well-separated clusters with significant inter-class distances. In contrast, the ICA-512 manifold exhibited reduced class spacing and more dispersed clusters. While FC100 proved significantly more challenging due to substantial class overlap, aligning with our global accuracy results, ICA-512 mitigated this overlap and improved class separation.

\section{Discussion}
The results across four diverse benchmarks consistently support our core hypothesis: high-capacity, off-the-shelf representations from models like DINOv2-L contain a highly generalizable and universal semantic manifold that is robust enough to bypass complex meta-learning optimization. Our discussion focuses on five key insights derived from the experimental data.

\textbf{Feature Maturation and the Semantic Plateau}
Our layer-wise characterization (Fig. \ref{fig:layer_effects}) reveals that feature discriminability follows a predictable sigmoidal progression. The high $R^2$ values for the logistic fits suggest that semantic maturation is not a linear process of depth, but rather a specialized emergence occurring in the final quarter of the network. The saturation observed at layers 21--24 (with the 24th layer being the norm) indicates a semantic plateau where the model has reached peak abstraction. This finding simplifies future deployments, as it suggests that researchers can reliably target these penultimate layers for fixed-feature extraction for these types of datasets.

\textbf{Dimensionality Reduction as Manifold Refinement}
One of the most significant findings is the performance of PCA-compressed embeddings in 1-shot settings. As seen in Table \ref{tab:fsl_imgnet_results} and Table \ref{tab:cifar_fc100_results}, projecting the 1024-dimensional raw features into 512 or 256 components can yield superior 1-shot accuracy (e.g., 84.16\% on miniImageNet). We attribute this to a regularizing effect: in the extreme low-data regime, high-dimensional latent spaces may contain idiosyncratic noise to which kNN classification is sensitive. By identifying the axes of maximum variance, PCA acts as a manifold refinement step that filters out low-variance noise while preserving the essential semantic signal.

\textbf{The Disentanglement Advantage of ICA}
While PCA dominated the 1-shot scenarios, the performance of ICA on the FC100 5-shot task is particularly noteworthy. Our ICA-512 variant achieved 78.74\%, nearly 10\% higher than previous state-of-the-art. FC100 is known for its fine-grained nature and low resolution, making class separation difficult. Unlike PCA, which maximizes variance, ICA seeks statistically independent components. This objective appears better suited for disentangling subtle, semantic features that distinguish closely related classes once the support set provides enough data (5-shot) to estimate the independent sources.

\textbf{Performance vs. Complexity}
Our framework highlights a significant efficiency gap in current few-shot research. By using a simple non-parametric kNN classifier on top of universal features, we exceeded the performance of complex architectures like SYNTRANS and SemFew by significant margins (e.g., +6.5\% on miniImageNet 5-shot). This suggests that the performance cliff for compression (Fig. \ref{fig:pca_effects}) is much lower than previously assumed; a representation can be reduced by 75\% (to 256 components) while still outperforming semantically informed meta-learning models. This shift toward "Good Embeddings" significantly lowers the barrier for deploying few-shot systems in resource-constrained environments.

\clearpage

\textbf{Future Outlook: Distillation of the Ideal Embedding Space.}
The performance gap between our DINOv2-L frozen baseline and contemporary few-shot architectures strongly supports the Platonic Representation Hypothesis that large architectures are converging on an "ideal" representation space and the "Rethinking Few-Shot" hypothesis that a sufficiently robust embedding is the most critical factor in few-shot success \cite{rethinking2020}. Our results demonstrate that high-capacity foundation models contain a highly generalizable semantic manifold that already outperforms sophisticated meta-learning algorithms.

Consequently, the next logical progression for the field is the structured distillation of this "ideal" embedding space into more efficient backbones. Our manifold refinement results, showing that accuracy actually increases when reducing the feature space to 256 or 512 components, provide a high-fidelity, low-noise target for this process. Future research should prioritize distillation paradigms that capture this universal logic within smaller architectures. Ultimately, this path reconciles the representational power of foundation models with the practical efficiency requirements of real-world few-shot systems.

\section{Conclusion} 
In conclusion, this paper has demonstrated that the universal latent manifold provided by high-capacity, off-the-shelf representations is inherently sufficient for state-of-the-art few-shot classification. By systematically mapping the hierarchical maturation of features within a frozen DINOv2 backbone, we identified a semantic plateau in the deepest layers that effectively bypasses the need for intensive backpropagation or complex meta-learning algorithms. Our results on four major benchmarks (miniImageNet, tieredImageNet, CIFAR-FS, and FC100) consistently show that a simple kNN classifier paired with robust embeddings can outperform sophisticated paradigms by significant margins, reaching accuracies as high as 96.51\% in 5-shot scenarios.

Furthermore, we established that linear manifold refinement through PCA and ICA provides an advantageous regularizing effect, particularly in 1-shot settings where dimensionality reduction filters idiosyncratic noise to improve generalization. These findings validate the Platonic Representation Hypothesis in the context of few-shot learning, suggesting that the bottleneck for rapid adaptation lies in the quality of the pre-trained representation rather than the complexity of the task-specific algorithm. By shifting the focus toward "Good Enough" embeddings, we provide a pathway for highly efficient, human-interpretable, and rapidly deployable recognition systems.

\par\vfill\par

\clearpage  % TODO FINAL: This \clearpage needs to be removed from both review and camera-ready versions.

%\section*{Acknowledgements}
%The authors would like to acknowledge the use of Gemini (Google) for assistance in the synthesis, editorial refinement, and structural organization of this manuscript, and Claude Code (Anthropic) for support in the implementation and optimization of the experimental pipeline. The authors have thoroughly verified all AI-generated contributions and take full accountability for the integrity, accuracy, and overall context of the final work.

\bibliographystyle{splncs04}
\bibliography{main}

\clearpage

% ---------------------------------------------------------------
% Include basic ECCV package
%\usepackage[review,year=2026,ID=11205]{eccv}

% ---------------------------------------------------------------
% Other packages

% --- TITLE AND AUTHORS ---
\title{Rethinking the Good Enough Embedding for Easy Few-Shot Learning: Supplemental Material} 

\titlerunning{Rethinking the Good Enough Embedding (Supp.)}

\author{Michael Karnes\inst{1} \and
Alper Yilmaz\inst{1}}

\authorrunning{M.~Karnes et al.}

\institute{The Ohio State University, Columbus, OH 43210, USA\\
\email{{karnes.30, yilmaz.15\}@osu.edu}}}

\maketitle

% ---------------------------------------------------------------
\section{Extended Layer-wise Few-Shot Performance Analysis}
\label{sec:layer_analysis}
\begin{figure}[ht]
\centering
\includegraphics[width=0.9\linewidth]{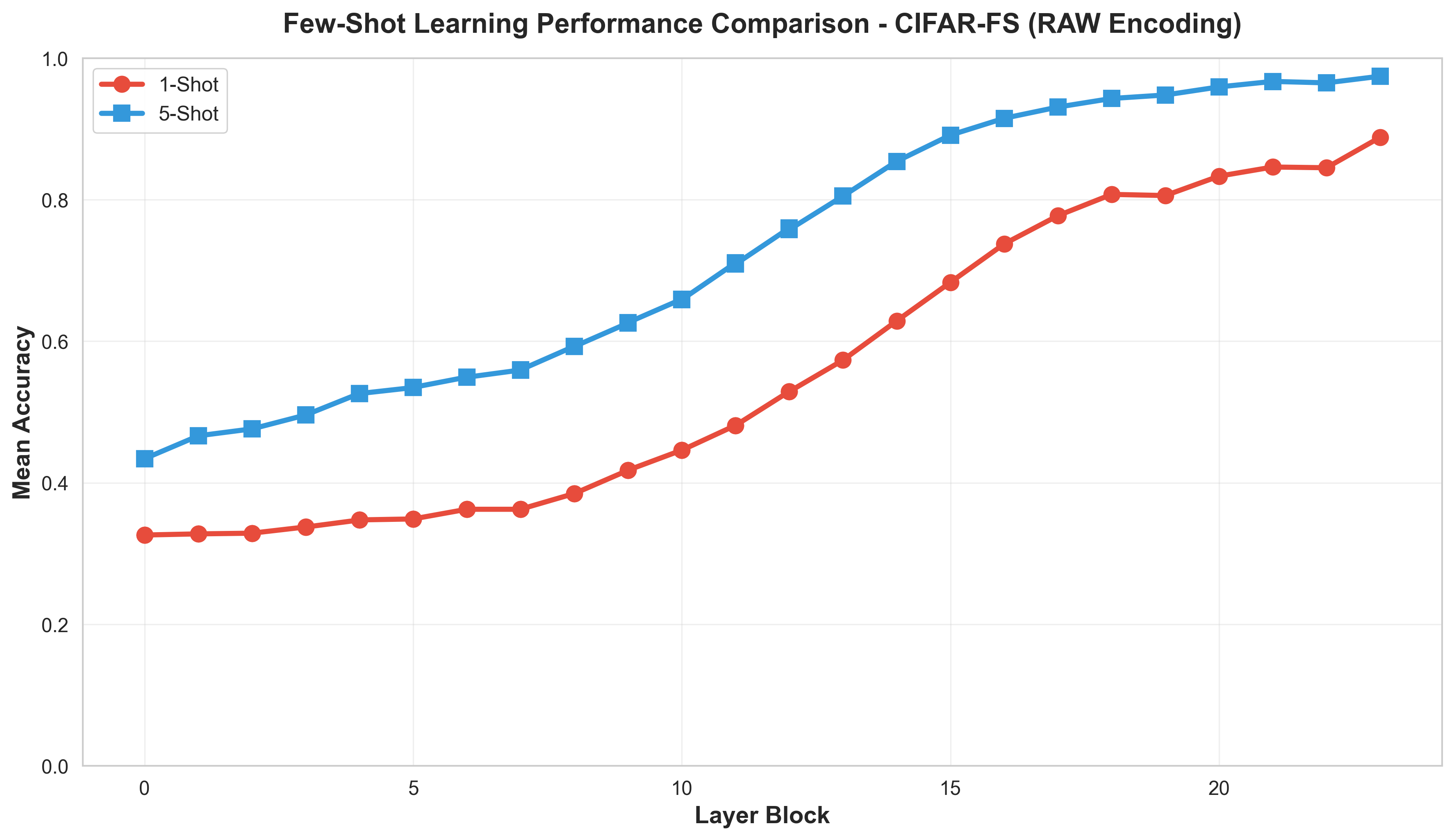}
\caption{Few-Shot Learning Performance Across Backbone Layers.
Mean accuracy are reported for 5-way tasks on the CIFAR-FS dataset across all 24 layer blocks of the frozen DINOv2-L backbone, comparing the performance of raw encodings in both 1-shot and 5-shot scenarios.}
\label{fig:cifar_FS_accuracy_progression}
\end{figure}

The fundamental premise of our work is that high-capacity, off-the-shelf representations contain an inherent "Good Embedding" that matures as it progresses through the hierarchical layers of a deep network. While the main manuscript identifies the optimal semantic plateau using a many-way characterization sweep (64 images per class across all training categories), this supplemental analysis provides a layer-wise characterization on the CIFAR-FS using a true few-shot (5-way) paradigm, evaluating both 1-shot and 5-shot regimes for every layer block in the DINOv2-L backbone. As illustrated in Figure \ref{fig:cifar_FS_accuracy_progression}, we observe a similar sigmoidal maturation process across both settings. Notably, the performance plateau for the 1-shot scenario occurs later than the 5-shot and continues to exhibit a marginal upward trend. This suggests that when support data is extremely scarce the most refined platonic features are particularly critical.

% ---------------------------------------------------------------

% ---------------------------------------------------------------

\section{Impact of Dimensionality Reduction}
\label{sec:dim_reduction}
\begin{figure}[htb]
\centering
\includegraphics[width=0.85\linewidth]{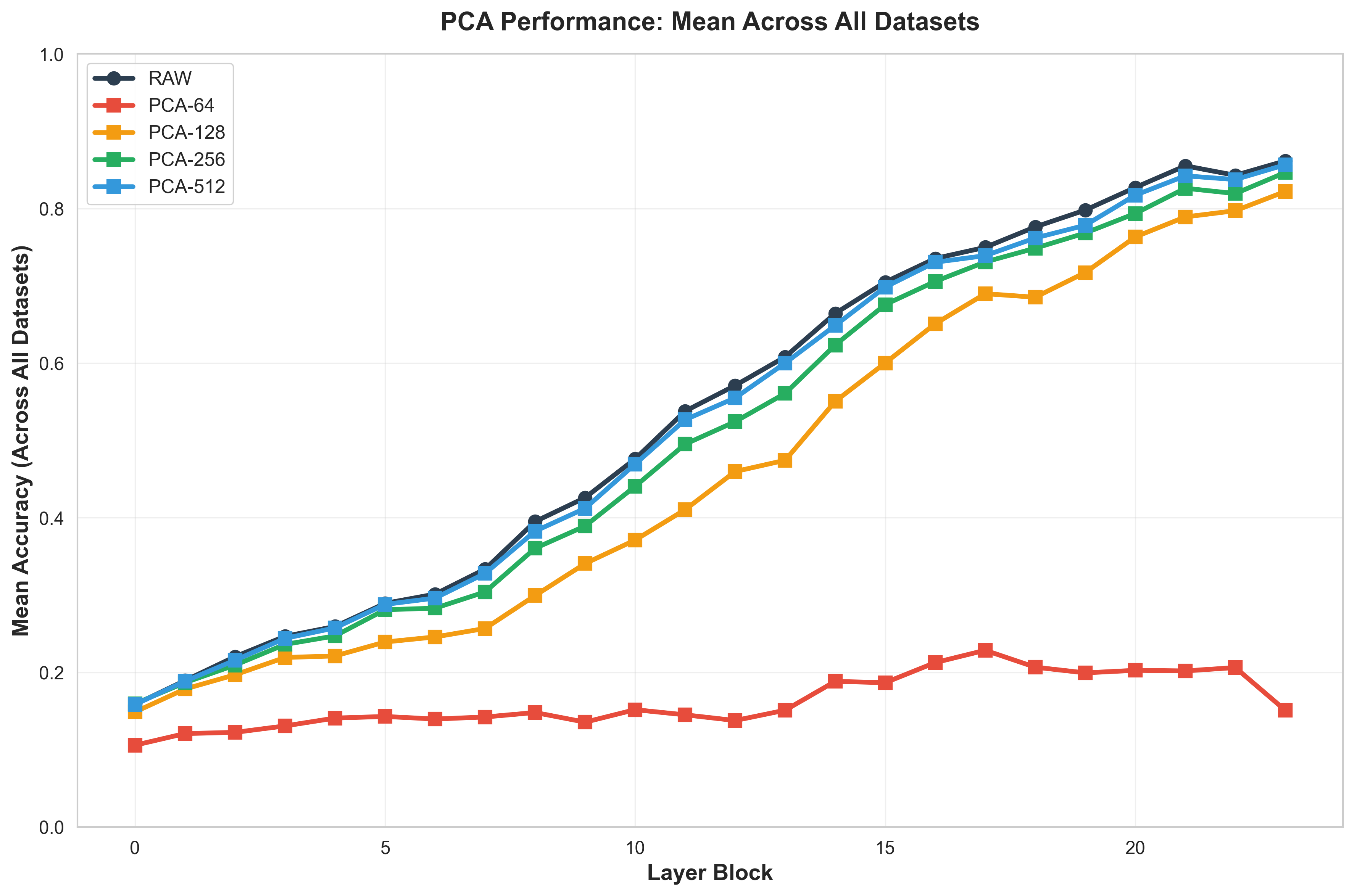}
\caption{Mean PCA performances aggregated across all four datasets using many-way classification (64 images per class). The results demonstrate that while the latent manifold is remarkably resilient to moderate compression (512 to 256 components), a significant performance decline exists at 64 components.}
\label{fig:pca_mean_progression}
\end{figure}

We further investigate the sensitivity of the latent manifold to different levels of dimensionality reduction using a many-way characterization protocol. For this analysis, we utilize 64 images per class from the training splits of each dataset to perform multi-class classification across all available categories. Figure~\ref{fig:pca_mean_progression} provides a comprehensive view of how classification performance scales from raw high-dimensional features (1024-D) down to highly compressed 64-component representations across the entire hierarchical depth of the backbone, averaged across the four evaluated datasets. Notably, as dimensionality is reduced, the characteristic sigmoidal progression of the logistic maturation curve becomes increasingly attenuated, eventually disappearing under 64-D compression. 

\begin{figure}[htb]
\centering
\includegraphics[width=0.85\linewidth]{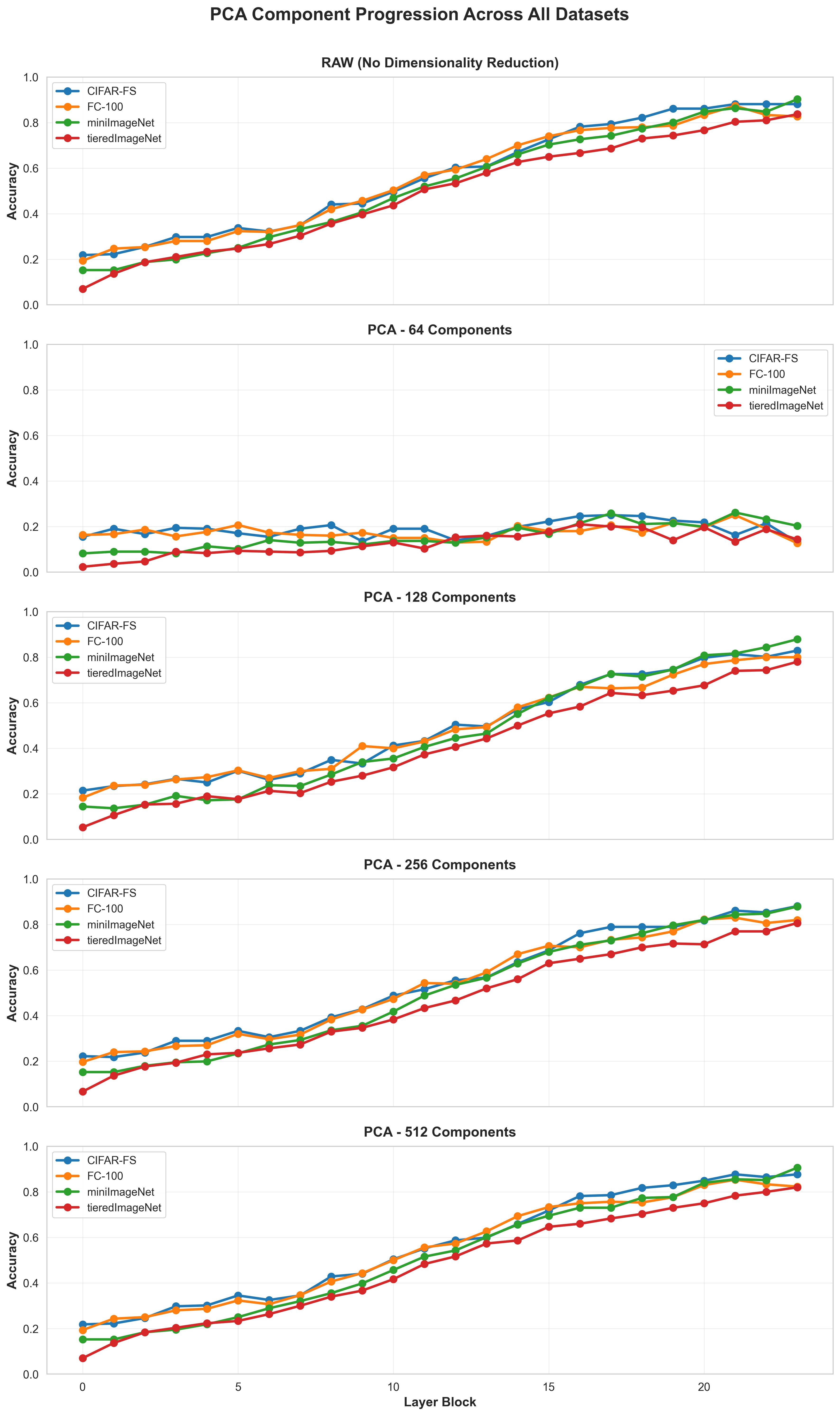}
\caption{Detailed PCA component progression for individual datasets (CIFAR-FS, FC100, miniImageNet, and tieredImageNet) under the many-way characterization setup. The sigmoidal maturation of features is preserved across 512, 256, and 128 dimensions, whereas the 64-component sweep collapses toward a near-random baseline, indicating a fundamental loss of semantic structure when using extreme compression.}
\label{fig:pca_component_progression}
\end{figure}

As illustrated in Figure \ref{fig:pca_component_progression}, the individual datasets exhibit remarkably synchronized architectural responses across the entire hierarchical depth of the backbone. Despite the significant differences in domain and granularity between CIFAR-FS, FC100, miniImageNet, and tieredImageNet, their performance trajectories under varying levels of PCA compression are nearly indistinguishable. This uniform behavior suggests that the "Good Enough Embedding" is not a dataset-specific phenomenon, but a fundamental property of the pre-trained backbone’s latent manifold.

% ---------------------------------------------------------------

\clearpage
% ---------------------------------------------------------------
\section{Manifold Visualization (t-SNE)}
\label{sec:tsne}
\begin{figure}[htb]
\centering
\includegraphics[width=0.5\linewidth]{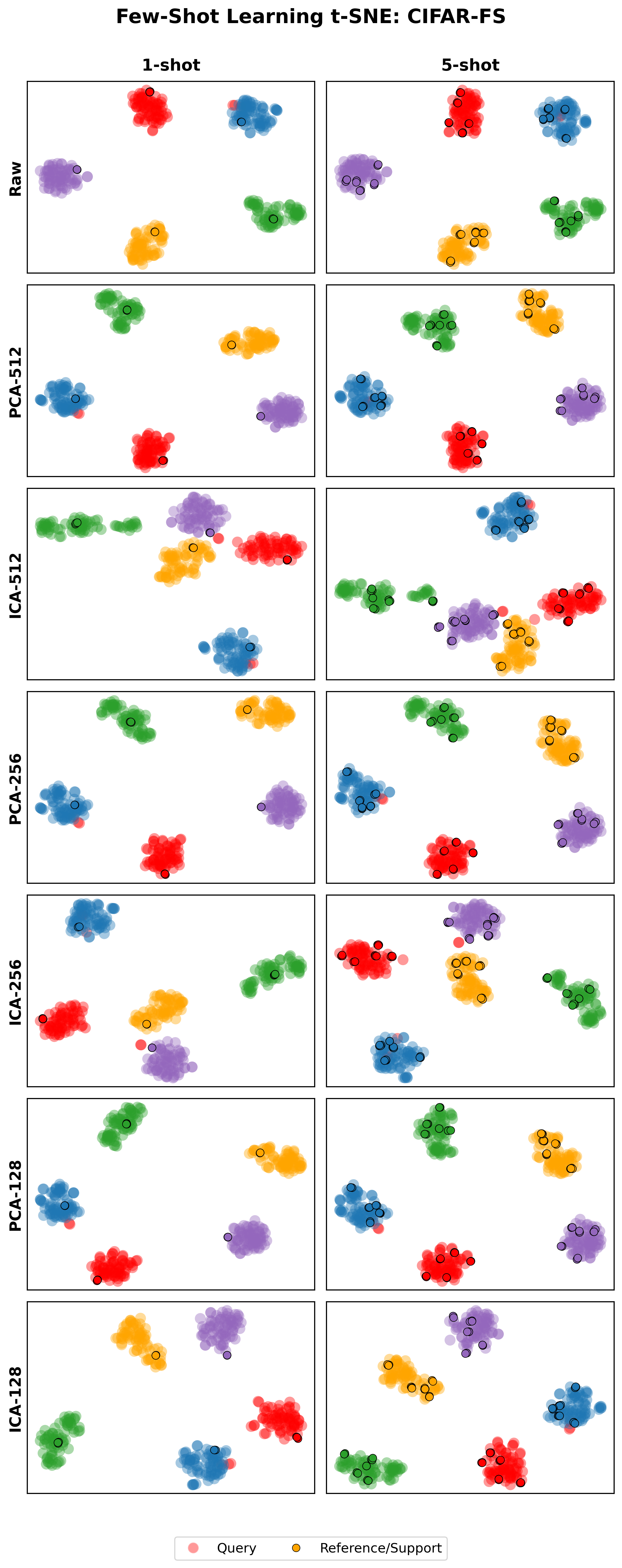}
\caption{t-SNE manifold comparison for CIFAR-FS. The clusters remain well-separated across all PCA and ICA reduction levels, with highly cohesive query-to-support mapping.}
\label{fig:cifarfs_tsne}
\end{figure}

As illustrated in Figures~\ref{fig:cifarfs_tsne}--\ref{fig:tieredimagenet_tsne}, the latent manifolds generated from the frozen DINOv2-L backbone exhibit remarkably distinct semantic clusters even in the absence of task-specific fine-tuning, with clear class separation visible across CIFAR-FS, miniImageNet, and tieredImageNet. The exception is the more challenging FC-100 dataset, where the clusters appear significantly more entangled, mirroring the lower quantitative accuracy reported for this benchmark. These qualitative results provide a visual sanity check for our state-of-the-art accuracy performance, demonstrating that while the "Good Enough Embedding" is inherently capable of resolving novel class boundaries, the complexity of the visual domain (e.g., the fine-grained nature of FC-100) directly affects how well the clusters separate in the latent space.

We observe a clear divergence in how PCA and ICA refine the latent space based on the underlying dataset complexity. On the easier benchmarks, such as CIFAR-FS and miniImageNet, PCA consistently maintains tighter and more evenly spaced clusters compared to ICA. In these regimes, the global variance maximization of PCA acts as an effective regularizer, preserving the intrinsic geometric order of the Platonic features with minimal distortion.

Conversely, in the more complex and fine-grained domains of FC-100 and tieredImageNet, ICA demonstrates superior local refinement. While the inter-class distances remain comparable to PCA, ICA tends to collapse the individual clusters into more unimodal, compact distributions. By seeking statistically independent components rather than just orthogonal ones, ICA appears to better disentangle the overlapping semantic signatures present in these difficult datasets, reducing the internal spread of the classes even when they remain in close proximity to one another.

\begin{figure}[htb]
\centering
\includegraphics[width=0.5\linewidth]{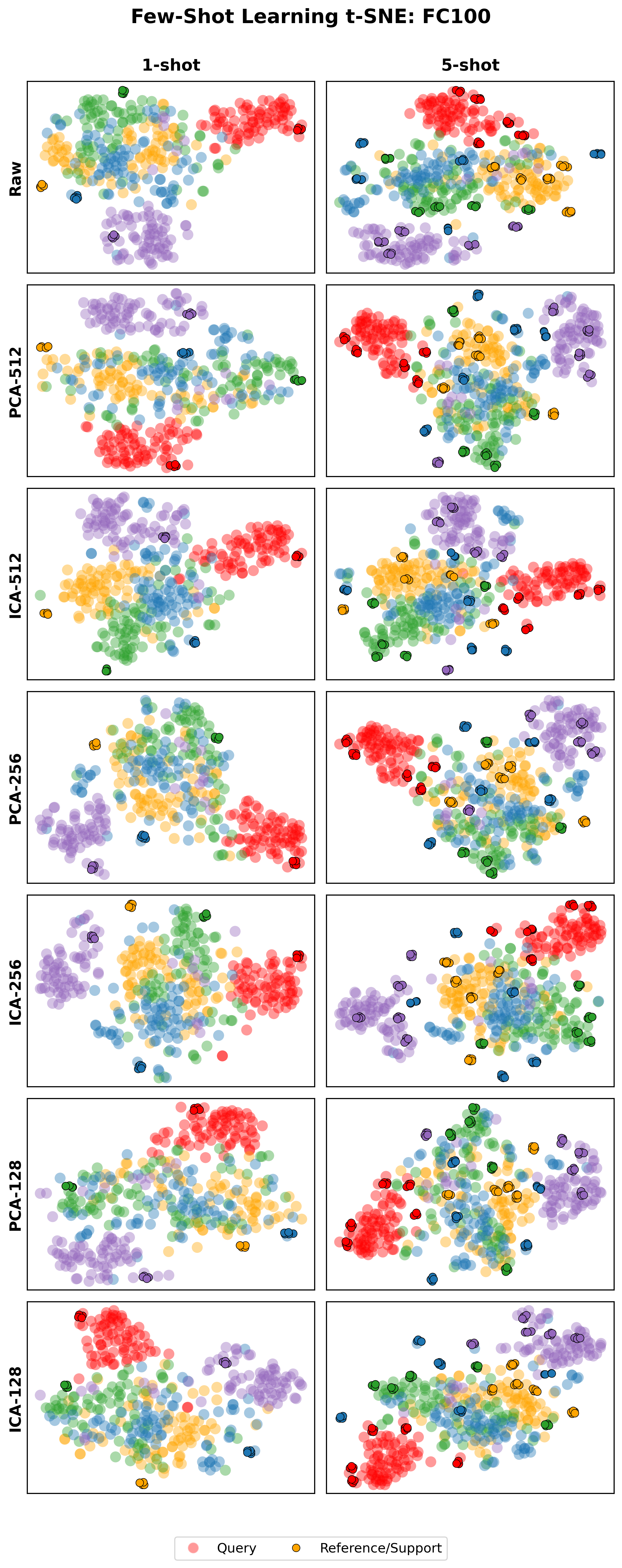}
\caption{t-SNE manifold comparison for the challenging FC100 dataset. Despite substantial class overlap in the raw latent space, the 5-shot ICA refinement notably improves inter-class discriminability.}
\label{fig:fc100_tsne}
\end{figure}

\begin{figure}[htb]
\centering
\includegraphics[width=0.5\linewidth]{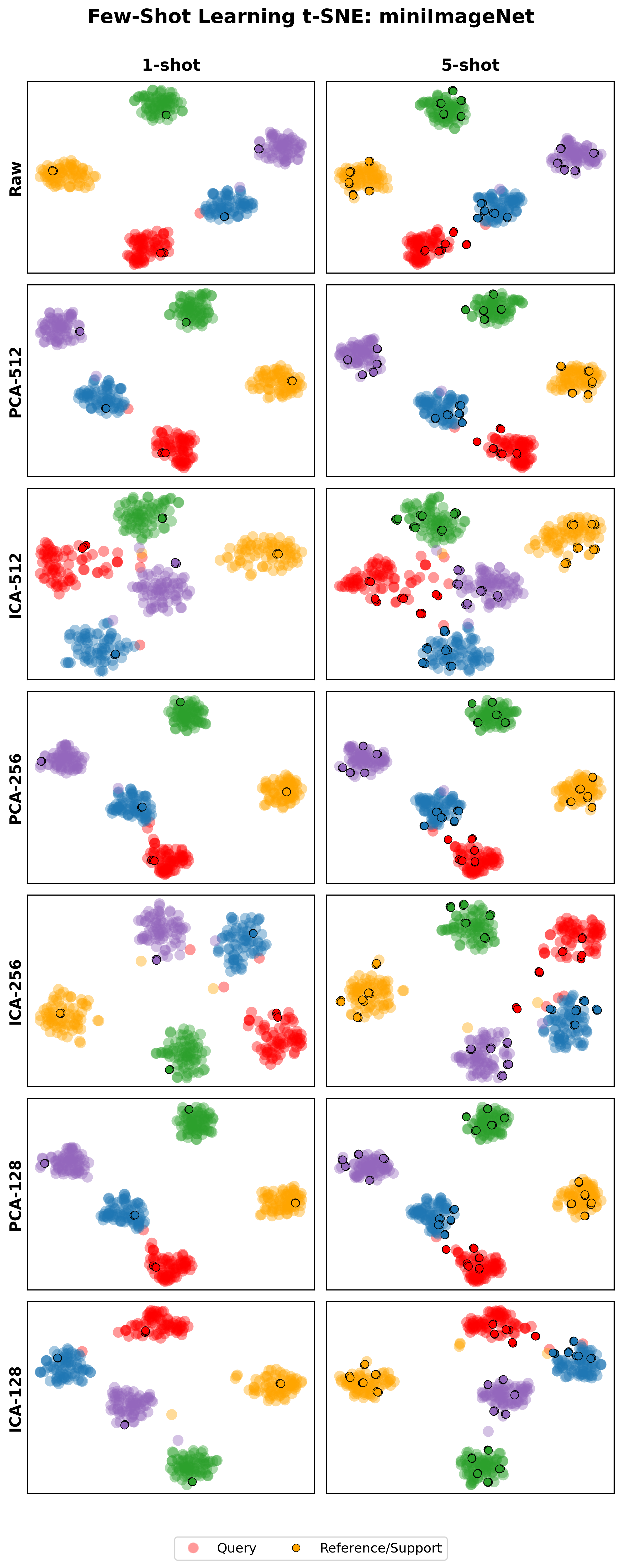}
\caption{t-SNE manifold comparison for miniImageNet. The manifold exhibits exceptional resilience to linear projection, maintaining high cluster purity even at 128 components.}
\label{fig:miniimagenet_tsne}
\end{figure}

\begin{figure}[htb]
\centering
\includegraphics[width=0.5\linewidth]{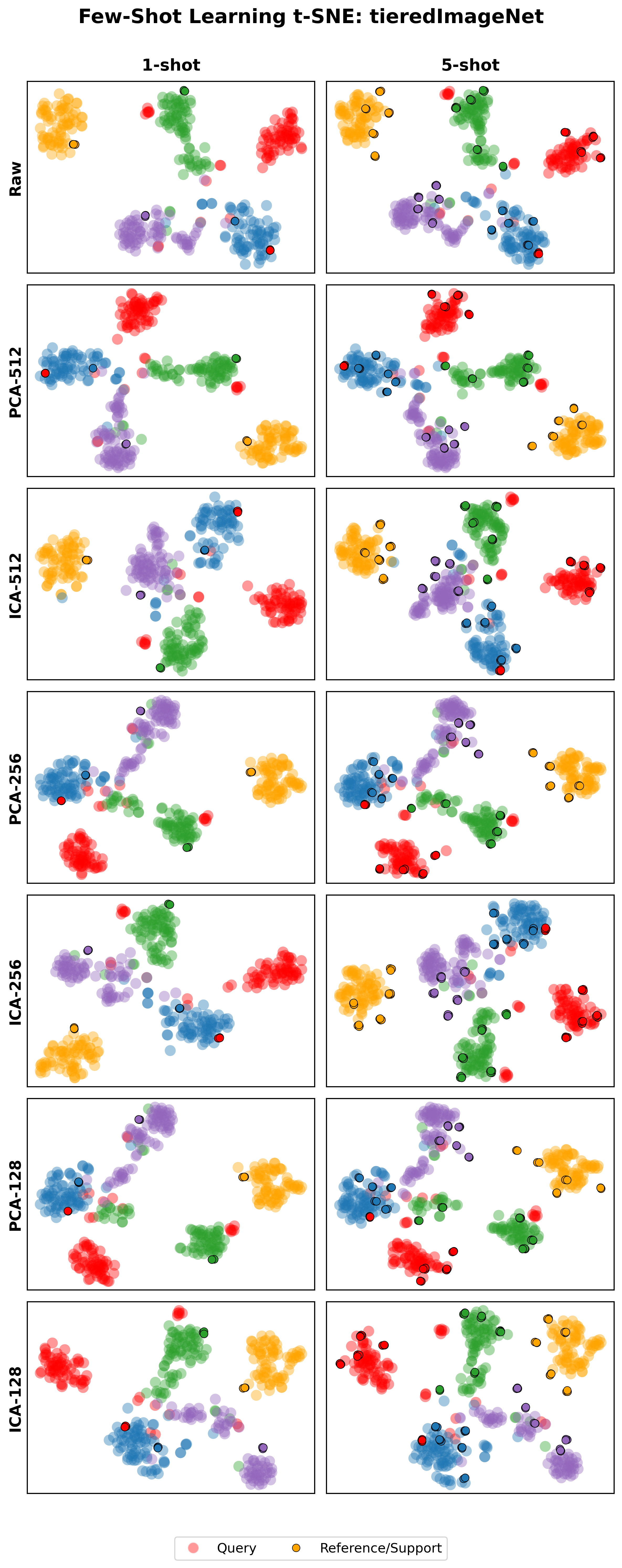}
\caption{t-SNE manifold comparison for tieredImageNet. The visualizations reinforce the observation that the Platonic feature structure is a robust, cross-domain backbone property.}
\label{fig:tieredimagenet_tsne}
\end{figure}

\clearpage
\section{Tabulated Layer-wise Accuracy Trends}
\label{sec:layerwise_tables}

\begin{figure}[htb]
\centering
\includegraphics[width=0.7\linewidth]{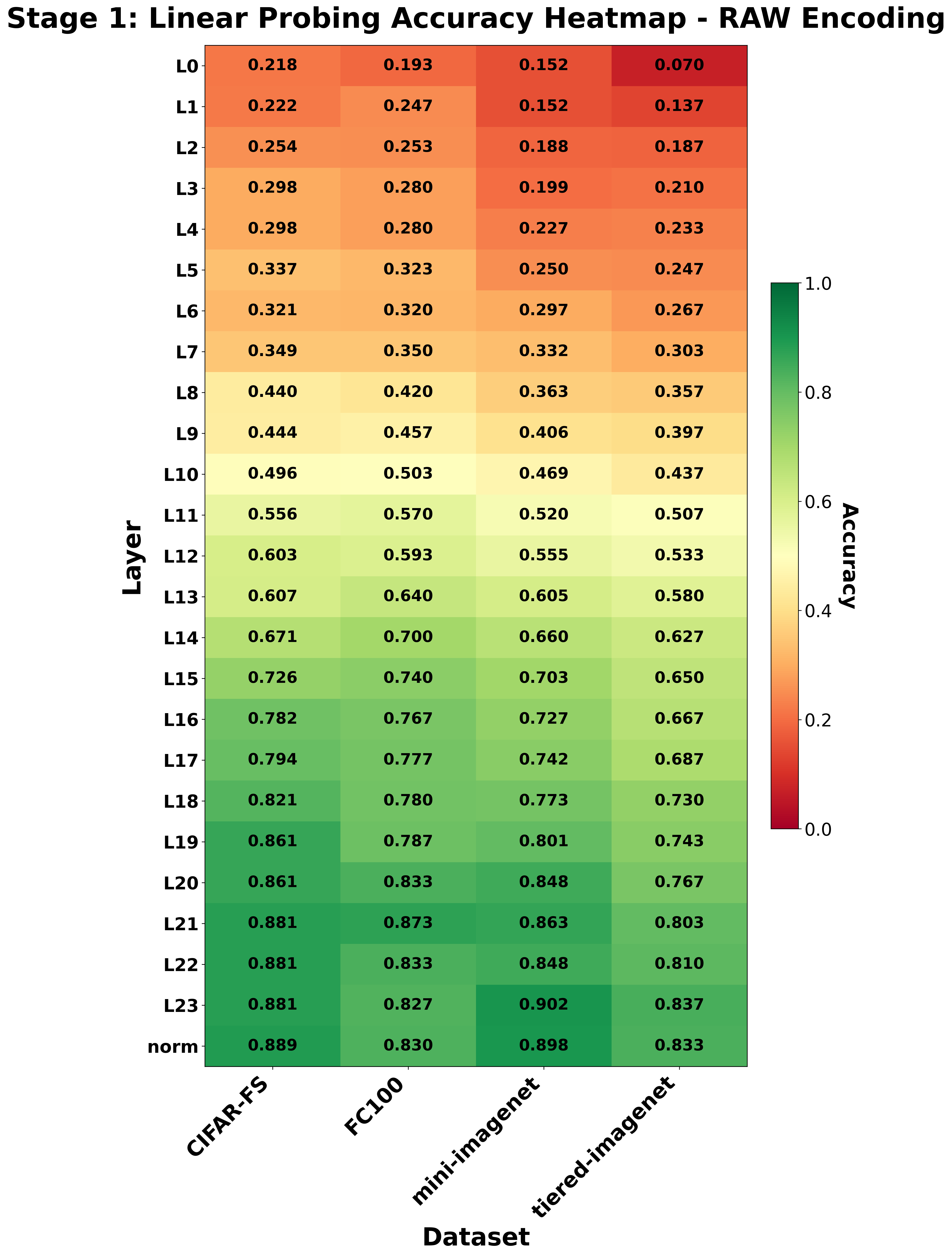}
\caption{This table provides the granular numerical results for the many-way characterization across all 24 layers (shown in Figure 2 in the manuscript). Colors indicate the accuracy gradient, highlighting the synchronized maturation of features as they progress toward the final layer and the post-backbone normalization.}
\label{fig:tieredimagenet_tsne}
\end{figure}

\begin{figure}[htb]
\centering
\includegraphics[width=0.7\linewidth]{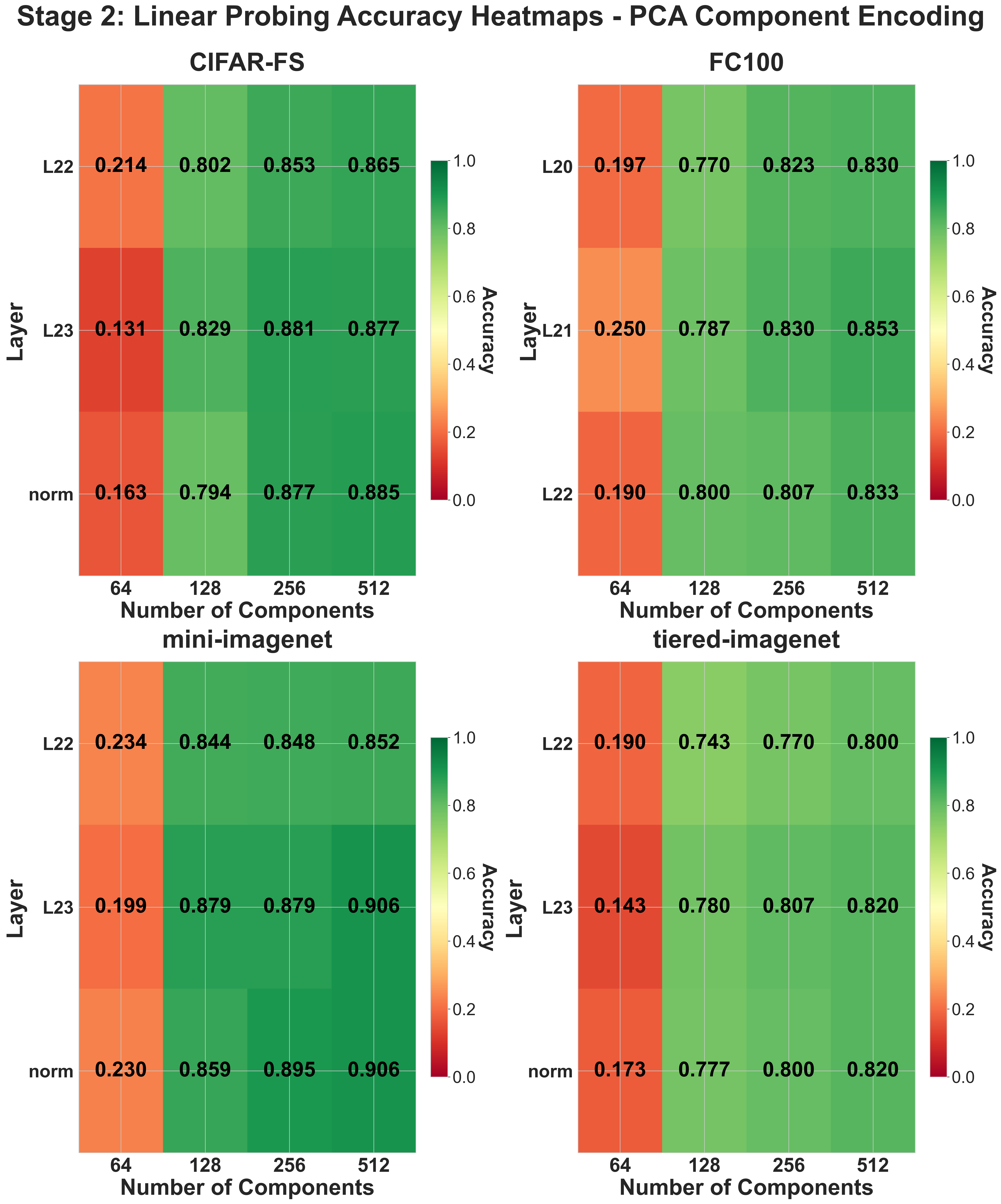}
\caption{A comparative numerical view of the optimal layer blocks and neighbors across four compression levels (shown in Figure 3 in the manuscript). The sharp color contrast at the 64 components provides quantitative evidence for the minimum intrinsic dimensionality required to sustain the latent manifold's discriminative structure.}
\label{fig:tieredimagenet_tsne}
\end{figure}

\clearpage
\section{Comparison across Backbones and Similarity Metrics}
\begin{figure}[htb]
\centering
\includegraphics[width=0.7\linewidth]{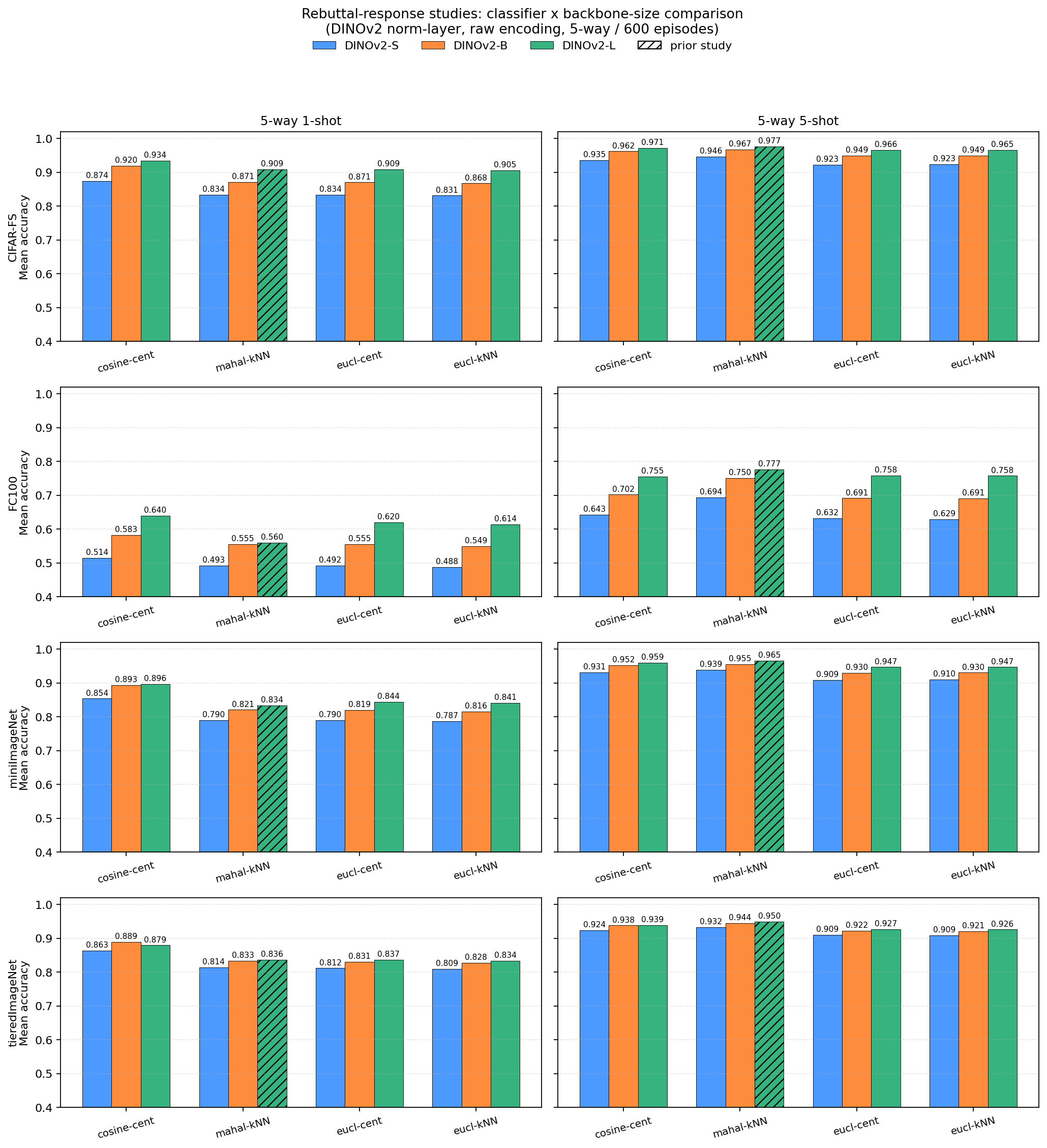}
\caption{A comparison of centroid of k-NN versus centroid and Mahalanobis, Euclidean, and Cosine similarity metrics across the four considered datasets. The Mahalanobis Distance with k-NN classification provided the highest accuracies for all 5-way 5-shot scenarios. The Cosine Similarity with centroids provided the highest accuracies for the 5-way 1-shot scenarios. The most significant effect of backbone selection appeared in the most challenging FC100 dataset and generally had larger impacts in the 5-way 1-shot scenarios. The effects of backbone selection diminish to generally incremental improvements in the 5-way 5-shot scenario.}
\label{fig:tieredimagenet_tsne}
\end{figure}

% ---------------------------------------------------------------

\end{document}